\documentclass{article}

\usepackage{arxiv}

\usepackage[utf8]{inputenc} 
\usepackage[T1]{fontenc}    
\usepackage{hyperref}       
\usepackage{url}            
\usepackage{booktabs}       
\usepackage{amsfonts}       
\usepackage{nicefrac}       
\usepackage{microtype}      
\usepackage{lipsum}
\usepackage{epsfig}  
\usepackage{amsmath}
\usepackage{algorithm}
\usepackage{algorithmic}   
\usepackage{graphicx}
\usepackage{color}
\usepackage[dvipsnames]{xcolor}
\usepackage{amssymb}  
\usepackage[algo2e]{algorithm2e}
\usepackage{tikz}
\usepackage{cite}
\usepackage{array}
\usepackage{romannum}
\usetikzlibrary{matrix,chains,positioning,decorations.pathreplacing,arrows}

\title{Data Driven Control with Learned Dynamics: Model-Based versus Model-Free Approach}

\author{
    Wenjian Hao \\
      Graduate Researcher\\
      Department of Mechanical Engineering\\
      Clemson University\\
      Clemson, SC 29634 USA \\
      \texttt{whao@g.clemson.edu} \\
    \And
    Yiqiang Han$^{*}$ \\
      Research Assistant Professor\\
      Department of Mechanical Engineering\\
      Clemson University\\
      Clemson, SC 29634 USA \\
      \texttt{yiqianh@g.clemson.edu}
    }

\begin{document}
\maketitle
\begin{abstract}
This paper compares two different types of data-driven control methods, representing model-based and model-free approaches. One is a recently proposed method – Deep Koopman Representation for Control (DKRC), which utilizes a deep neural network to map an unknown nonlinear dynamical system to a high-dimensional linear system, which allows for employing state-of-the-art control strategy. The other one is a classic model-free control method based on an actor-critic architecture – Deep Deterministic Policy Gradient (DDPG), which has been proved to be effective in various dynamical systems. The comparison is carried out in OpenAI Gym, which provides multiple control environments for benchmark purposes. Two examples are provided for comparison, i.e., classic Inverted Pendulum and Lunar Lander Continuous Control. From the results of the experiments, we compare these two methods in terms of control strategies and the effectiveness under various initialization conditions. We also examine the learned dynamic model from DKRC with the analytical model derived from the Euler-Lagrange Linearization method, which demonstrates the accuracy in the learned model for unknown dynamics from a data-driven sample-efficient approach.
\end{abstract}


\section{Introduction}

The Koopman operator is an infinite-dimensional operator that governs the evolution of scalar observations in the state space for a nonlinear dynamical system, as shown in Equation \ref{eqn.koopman}. It has been proven \cite{koopman1931hamiltonian} that there exists a Koopman Operator $\mathcal{K}$ that can advance measurement functions in a linear operator fashion in an infinite-dimensional space.
\begin{equation}\label{eqn.koopman}
    \begin{gathered}
        \mathcal{K} g = g \circ \boldsymbol{F}
    \end{gathered}
 \end{equation}
where, the $\boldsymbol{F}$ represents the dynamics that map the state of the system forward in time; the $g$ is the representation of the system in the infinite-dimensional space. For practical use, existing approaches attempt to compute a finite-dimensional approximation of this operator. Dynamic Mode Decomposition (DMD) has been proven to be an effective way to find reduced-order models to represent the higher-dimensional complex systems \cite{williams2015data, williams2016extending, williams2014kernel, koopman1931hamiltonian}. For more extension work of Koopman operator methods to controlled dynamical refer to \cite{korda2018linear, proctor2016dynamic, kaiser2017data, kaiser2020data, broad2018learning, you2018deep,ma2019optimal}. However, the use of the Koopman operator as a linear predictor has been hindered by the computational complexity. A major concern is that, as the dimension of state space increases, the numerical method to approximate Koopman operator is running at a rapidly increasing polynomial time complexity to compute a rich set of basis functions. Another route to find Koopman operator approximation is through using Deep Neural Networks (DNN). The DNN provides a tremendous capacity to store and map lower-dimensional measurements to higher-dimensional lifted representations. Fruitful research results have been found in applying DNN in the linear embedding of nonlinear dynamics field \cite{lusch2018deep}.

Deep Learning of Koopman Representation for Control (DKRC) exemplifies a recent research trend of using DNN as a linear embedding tool to learn nonlinear dynamics and control the system based on the learned dynamics. In this framework, the optimal control is being achieved by combining a global model of the system dynamics and a local model for local replanning. The key feature of DKRC is that it extends the use of DNN for the finite-dimensional representation of the Koopman operator in a controlled dynamical system setting. It benefits from a data-driven learning algorithm that can automatically search for a rich set of suitable basis functions ($\psi(x)$) to construct the approximated linear model in the lifted space \cite{Han:2020f}. We can then rewrite the Equation \ref{eqn.koopman} into Equation \ref{eqn.koopman_psi}:
\begin{equation}\label{eqn.koopman_psi}
    \begin{gathered}
        \psi(x_{t+1}) = \mathcal{K} \psi(x_t)
    \end{gathered}
 \end{equation}

In this paper, we implement the Model Predictive Control (MPC) after learning the dynamics of the lifted high-dimensional linear system with DKRC. Other methods, such as regression or DAGGER (i-LQR or MCTS for planning) methods have been introduced after the learned model is available. However, those methods suffer from distribution mismatch problems or issues with the performance of open-loop control in stochastic domains. We propose using MPC iteratively, which can provide robustness to small model errors that benefit from the close loop control. As authors mentioned in Ref \cite{psrl}, if the system dynamics were to be linear and the cost function convex, the global optimization problem solved by MPC methods would be convex, which means in this situation the convex programming optimization techniques could be promising to achieve some theoretical guarantees of convergence to an optimal solution. Therefore MPC would certainly perform better in a linear world. MPC represents state of the art for the practice of real-time optimal control \cite{ddmpc}. MPC is an online optimization solver method that seeks optimal state solutions at each time step under certain defined constraints. We only take the first value of the planned control sequence of each time step. MPC has been proven to be tolerant to the modeling errors within Adaptive Cruise Control(ACC) simulation \cite{acc} \cite{cprm}. To solve the convex problem, we used CVXPY \cite{cvxpy} in this study.

Deep Deterministic Policy Gradient (DDPG) is a model-free, off-policy algorithm using DNN function approximators that can learn policies in high-dimensional, continuous action spaces \cite{ddpgo}. DDPG is a combination of the deterministic policy gradient approach and insights from the success of Deep Q Network (DQN) \cite{dqno}. DQN is a policy optimization method that achieves optimal policy by inputting observation and updates policy by back-propagate policy gradients. However, DQN can only handle discrete and low-dimensional action spaces, limiting its application considering many control problems have high-dimensional observation space and continuous control requirements. The innovation of DDPG is that it extends the DQN to continuous control domain and higher dimensional system and has been claimed that it robustly solves more than 20 simulated physics tasks, including classic problems such as cartpole swing-up, dexterous manipulation, legged locomotion and car driving \cite{ddpgo}. In addition, DDPG has been demonstrated to be sample-efficient compared to the Covariance Matrix Adaptation Evolution Strategy (CMA-ES), which is also a black-box optimization method widely used for robot learning \cite{ddpgc}.

Since DKRC and DDPG are both capable of solving dynamic systems control problems with high-dimensional and continuous control space, direct comparison is desirable by constructing the two algorithms from the ground up and applying them to solve the same tasks. We also compare the DKRC to the analytical model obtained by the classic Euler-Lagrange Linearization method. The comparison are examined in the Inverted pendulum environment and Lunar Lander Continuous Control environment in OpenAI Gym \cite{brockman2016openai}. 

This paper is organized into four sections. In Section \Romannum{2}, we briefly introduce the algorithms we use to achieve the comparison results. In Section \Romannum{3}, we set up the optimization problems in our simulation environment. In Section \Romannum{4}, we present the comparison results between the model-based and model-free approaches. Section \Romannum{5} concludes. 

\section{Algorithm}
In this section we briefly introduce the algorithm of DDPG and DKRC.

\subsection{Deep Koopman Representation for Control}
Deep Koopman Representation for Control (DKRC) is a model-based control method which transfers a nonlinear system to a high-dimensional linear system with a neural network and deploys model-based control approaches like model predictive control (MPC). DKRC benefits from massive parameters of the neural network. By utilizing neural networks as a lift function, we can get a robust transformation from a nonlinear system to a lifted linear system as requirements. This paper also presents an auto decoder neural network \cite{dec} to map the planned states of MPC in lifted state-space back to the non-lifted state space to verify the DKRC control beyond the above comparison. A schematic diagram of the DKRC framework is shown in Figure \ref{fig:DKRC}:

\begin{figure}[ht]
    \centering
    \includegraphics[width=0.65\textwidth]{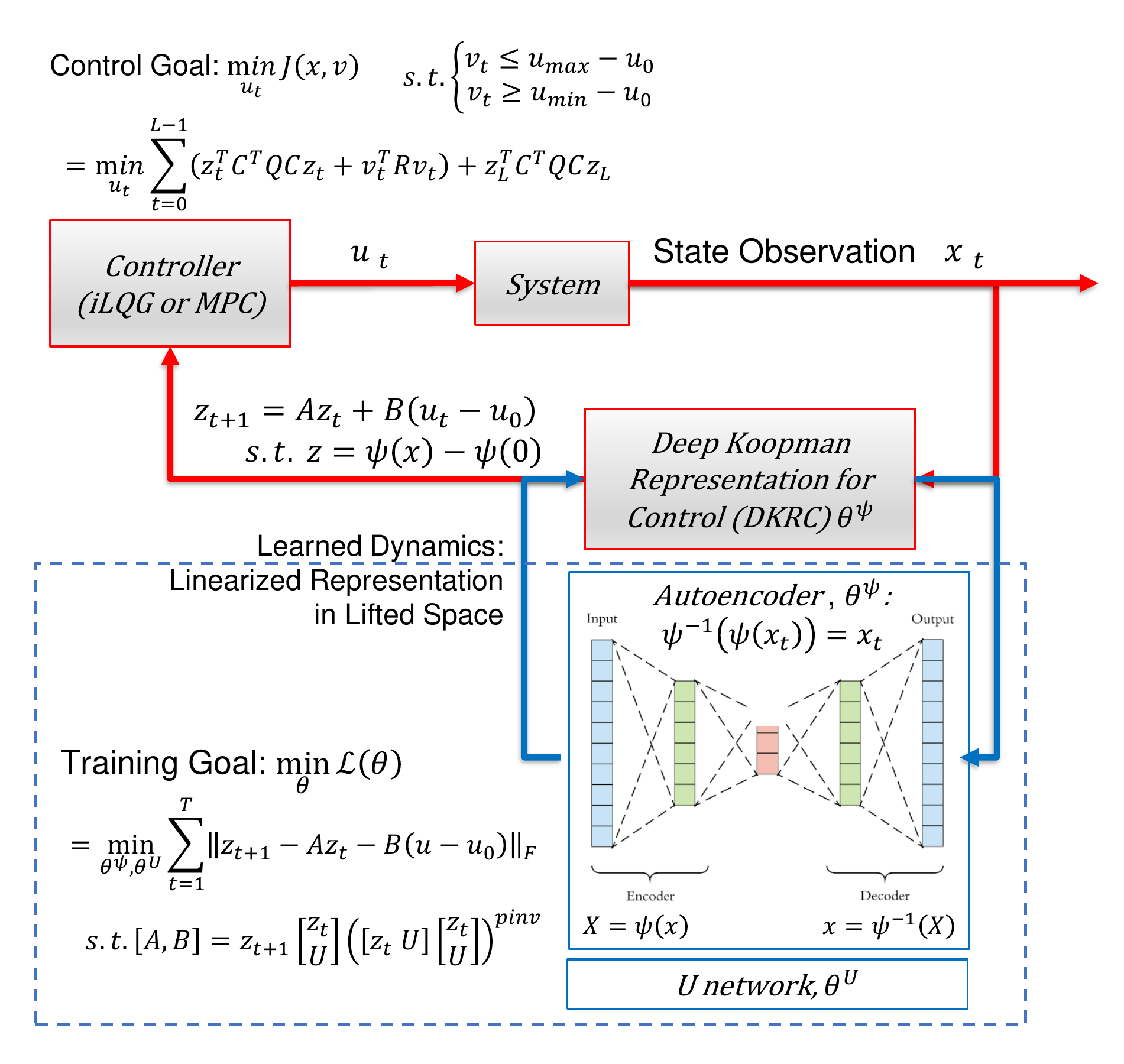}
    \caption{Schematics of DKRC framework}\label{fig:DKRC}
\end{figure}

The implementation of DKRC can be divided into the following steps: 
\begin{enumerate}
    \item 
    Build a neural network $\psi_{N}(x|\theta)$, where $N$ is the dimension we intend to lift, $x_t$ is the state space observations of the dynamical system at time $t$, $\theta^\psi$ represent parameters of the neural network. We decouple the optimal control problem into two function modules. The DKRC module handles the mapping process from the state space to the system in higher-dimensional space, which exhibits linear system behavior. We then feed the mapped states $z_t$ to the controller, which uses state-of-the-art planning algorithms to construct cost-to-go function and plan optimal control for the system. Those state-of-the-art control algorithms, such as iLQG and MPC, typically cannot work to the best efficiency when the system is highly nonlinear in nature. The introduction of the DKRC algorithm provides promising results in expanding the success of those conventional control algorithms into the field of challenging control problems in nonlinear dynamical systems.
    \item
    The DKRC algorithm achieves the lifting from low-dimension space to high-dimension space using the recent success of the deep neural networks in the control field. We use an autoencoder neural network structure as the backbone of DKRC. The encoder part of the autoencoder network will serve as the lifting function that maps the low-dimensional state observations to high-dimensional Koopman representations. Considering DKRC executes optimal MPC control in a lifted high-dimensional state space, it is also interesting to know how the DKRC makes control planning in high dimensional state space. To verify DKRC's planning behavior in a high-dimensional state space, we build a decoder neural network that maps the lifted space back to the original space. The schematic of the relationship between DKRC encoder and decoder is shown in Figure \ref{fig:decoder}. 
    
    In this example, we have three state-space observations, for which we lift to Koopman representation with a dimension of eight (intermediate result in the center hidden layer). The decoder network will also map the lifted states back to the original state space with a dimension of three as the resulting output. Through this way, we can verify that our approach is behaving consistently in mapping between the low and high dimensional spaces, which is vital in assessing its reliability and avoiding a complete black-box behavior from a simple feed-forward neural network without validation.
    
    \begin{figure}[ht]
    \centering
    \includegraphics[width=0.5\textwidth]{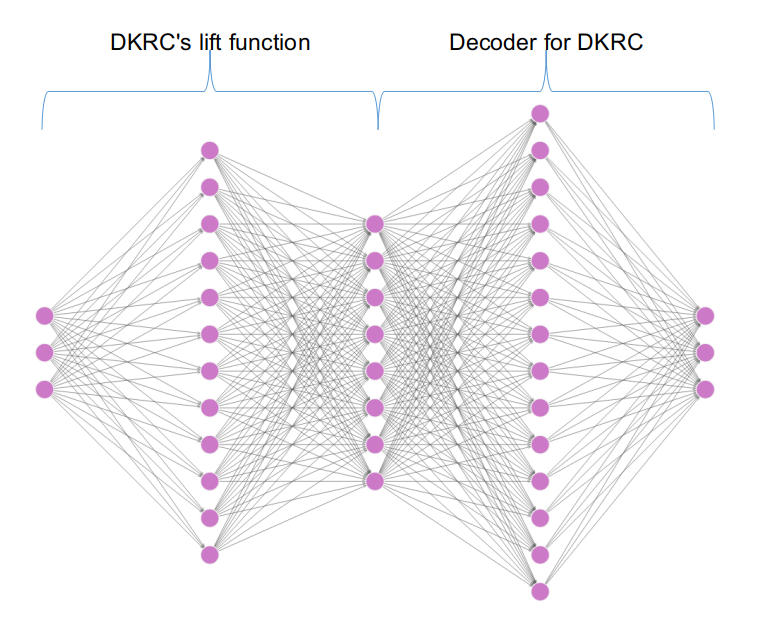}
    \caption{Schematics of autoencoder neural network sturcture of DKRC}\label{fig:decoder}
    \end{figure}
    
    \item
    To tune the parameters of the autoencoder network ($\theta^\psi$), we carefully design and closely monitor loss functions during the training process. Two loss functions are used in this task. We define the first loss function, $L1$, in Equation \ref{eqn.loss1},
    \begin{equation}\label{eqn.loss1}
        \begin{gathered}
         L1(\theta)=\frac{1}{L-1}\sum_{t=0}^{L-1}\parallel \psi(x_{t+1};\theta)-K*\psi(x_t;\theta)\parallel\\ K=\psi(x_{t+1};\theta)*\psi(x_t;\theta)^\dagger
        \end{gathered}
     \end{equation}
    \item
    The $\dagger$ sign represents the pseudo-inverse operation. By minimizing this loss function, we ensure that the Koopman operator theorem in Equation \ref{eqn.koopman} and \ref{eqn.koopman_psi} is being strictly enforced.
    Since we are interested in the dynamical system with control, we seek to identify the linearized system with $A$ and $B$ coefficient matrices. During the model training iterations, we can get the matrices by minimizing Equation \ref{eqn.ab}
     \begin{equation}\label{eqn.ab}
        \begin{gathered}
            M = [A,B] = argmin_{A,B}{|| \psi_{N}(X_{t+1}|\theta) - {A}\psi_{N}{({X_t}}|\theta) - {B}{U_t} ||}\\
        \end{gathered}
     \end{equation}
    For larger data sets with $K\gg N$, instead of solving the least square problem associated with Equation \ref{eqn.ab} directly, it is beneficial to solve a slightly modified normal equation as defined in Equation \ref{eqn.absol}
     \begin{equation}\label{eqn.absol}
        \begin{gathered}
            \boldsymbol{V} = M \boldsymbol{G}
        \end{gathered}
     \end{equation}
     where,\ 
         $$\boldsymbol{G} = {\begin{bmatrix}
            \psi_{N}(X_t|\theta)\\
            U_t\\\end{bmatrix}}{\begin{bmatrix}
            \psi_{N}(X_t|\theta)\\
            U_t\\\end{bmatrix}}^{T}, \ 
     \boldsymbol{V} = \psi_{N}(X_{t+1}|\theta){\begin{bmatrix}
            \psi_{N}(X_t|\theta)\\
            U_t\\\end{bmatrix}}^{T}$$\\
     Any solution to Equation \ref{eqn.absol} is a solution to Equation \ref{eqn.ab}. The size of the matrices $\boldsymbol{G}$ and $\boldsymbol{V}$ is the $(N+m)\times (N+m)$ and $N\times (N+m)$ respectively, hence independent of the number of samples ($K$) in the data set \cite{lpfnd}.\\
     We can also define a second loss function $L2(\theta)$ in Equation (\ref{eqn.loss2}), which ensures a controllable lifted linear states space.
    \begin{equation}\label{eqn.loss2}
        \begin{gathered}
        L2(\theta) = (N-rank(controllability(A,B)))
        \end{gathered}
    \end{equation}
     \item
     After training converges, we achieve a neural network model $\psi_{N}(x|\theta)$ that is capable of lifting state space to high-dimensional space with learned dynamic models coefficients $A$, $B$, and $C$. The $C$ matrix is necessary for later controller cost function design and is being defined as $C = X_t*\psi_{N}(X_t|\theta)^\dagger$
     \item
     Finally, we implement Model Predictive Control (MPC) or Linear-Quadratic Regulator (LQR), with a cost function in the format of $J(V) =\sum_t \psi_{N}(X_t|\theta)^\top C^\top Q C \psi_{N}(X_t|\theta)+u_t^\top R u_t$, \\ where $Q_{lifted} = C^{T} Q C$
    
\end{enumerate}

To summarize the above discussion and also the information depicted in Figure \ref{fig:DKRC}, the algorithm is listed below:
\DontPrintSemicolon
\begin{algorithm}[ht]
\SetAlgoLined
\KwIn{observations: x, control: u}
\KwOut{Planned trajectory and optimal control inputs: ($z_{plan}$, $v_{plan}$)}
\begin{itemize}
    \item Initialization 
        \begin{enumerate}
            \item Set goal position $x^*$
            \item Build Neural Network:
            $\psi_N(x_t;\theta)$
            \item Set $z(x_t;\theta) = \psi_N(x_t;\theta) - \psi_N(x^*;\theta)$
        \end{enumerate}
    \item Steps
    \begin{enumerate}
        \item Set $K=z(x_{t+1};\theta)*z(x_t;\theta)^\dagger$
        \item Set the first loss function $L1$\\
        $L1(\theta)=\frac{1}{L-1}\sum_{t=0}^{L-1}\parallel z(x_{t+1};\theta)-K*z(x_t;\theta)\parallel$
        \item Set the second loss function $L2$\\
        $[A,B]={\bf z}_{t+1}
        \begin{bmatrix}
            {\bf z}_{t} \\
               U 
         \end{bmatrix}
         \begin{pmatrix}
            \begin{bmatrix}
               {\bf z}_{t}\  
               U 
            \end{bmatrix}
            \begin{bmatrix}
               {\bf z}_{t} \\
               U 
            \end{bmatrix}
         \end{pmatrix}^\dagger$\\
        
         $L2(\theta) = (N-rank(controllability(A,B)))+||A||_1+||B||_1$
         \item Train the neural network, updating the complete loss function\\ $L(\theta)=L1(\theta)+L2(\theta)$
         \item After converging, We can get system identity matrices A, B, C\\
         $C=\boldsymbol X_t*\psi_N(\boldsymbol X_t)^\dagger$
         \item Apply LQR or MPC control with constraints
    \end{enumerate}
\end{itemize}
\caption{Deep Koopman Representation for Control (DKRC)}
\end{algorithm}

\subsection{Deep Deterministic Policy Gradient}
\begin{figure}[ht]
    \centering
    \includegraphics[width=0.65\textwidth]{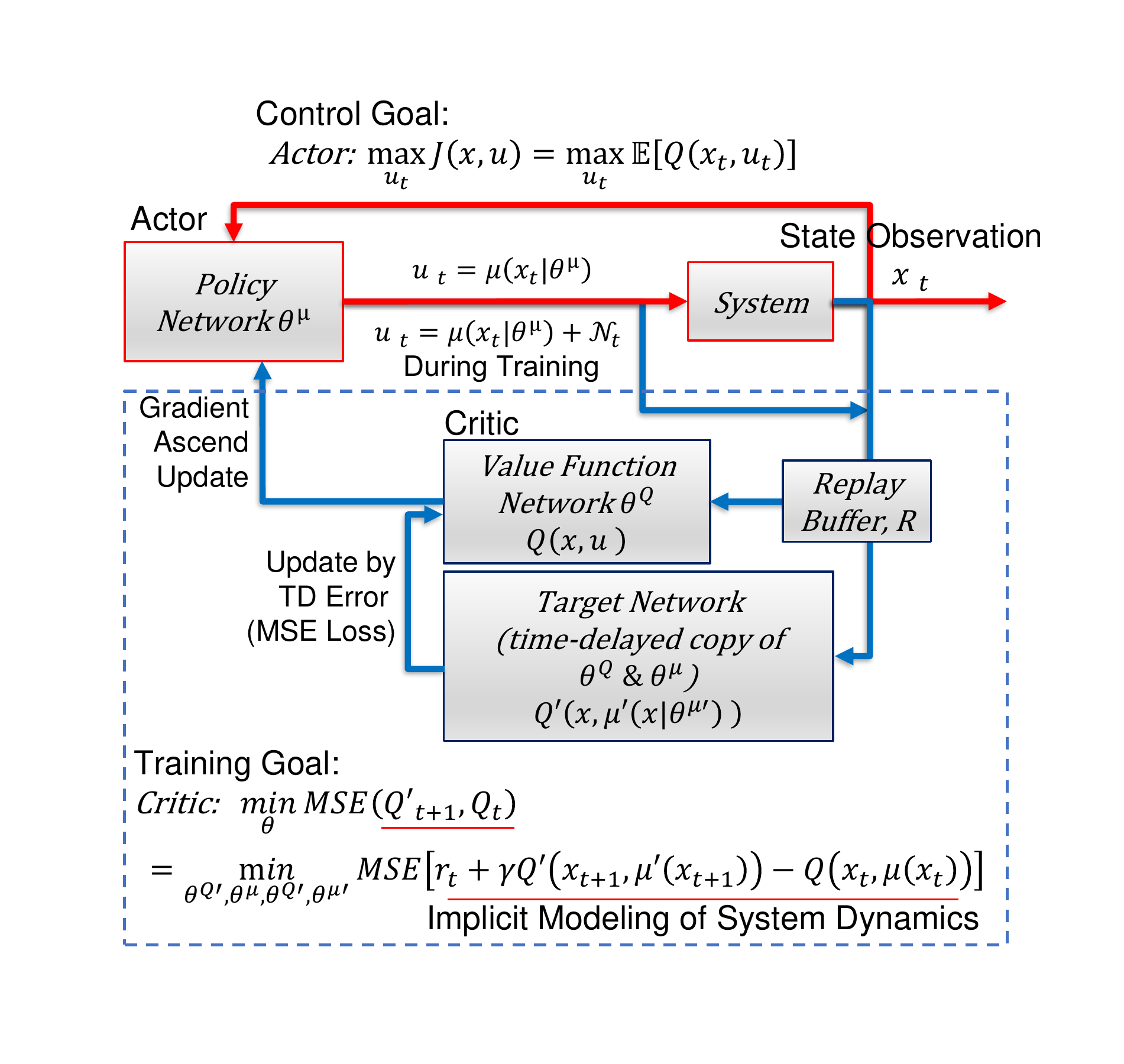}
    \caption{Schematics of DDPG framework}\label{fig:DDPG}
\end{figure}

Deep Deterministic Policy Gradient (DDPG) is a reinforcement learning algorithm based on an actor-critic framework. The DDPG and its variants have been demonstrated to be very successful in designing optimal continuous control for many dynamical systems. However, compared to the above model-based optimal control, DDPG employs a more black-box style neural network structure that outputs control directly based on learned DNN parameters. The DDPG and its variants typically require a significant amount of training data and a good design of the reward function to achieve good performance without learning an explicit model of the system dynamics. A schematic diagram of the DDPG framework is shown in Figure \ref{fig:DDPG}.

 The control goal of the DDPG algorithm is to find control action $u$ to maximize the rewards (Q-values) evaluated at the current time step. The training goal of the algorithm is to learn neural network parameters for the above networks so that rewards can be maximized at each sampled batch, while still being deterministic. The training of the DDPG network will then follow a gradient ascend update from each iteration, as defined in Equation \ref{eqn.DDPG_grad}

\begin{equation}\label{eqn.DDPG_grad}
    \begin{gathered}
        {\nabla_{\theta^{\mu}}}J \approx \mathbb{E}[\nabla_{\theta^{\mu}}Q(x,u|\theta^Q)|x=x_t, u=\mu(x_t|\theta^\mu)]
    \end{gathered}
\end{equation}

In Equation \ref{eqn.DDPG_grad}, we have two separate networks involved: Value function network $\theta^Q$ and policy network $\theta^\mu$. To approximate the gradient from those two networks, we design the training process relying on one replay buffer and four neural networks: Actor, Critic, Target Actor, Target Critic.

At the gathering training samples stage, we use a replay buffer to store samples. A replay buffer of DDPG essentially is a buffered list storing a stack of training samples $(x_t, u_t, r_t, x_{t+1})$. During training, samples can be drawn from the replay buffer in batches, which enables training using a batch normalization method \cite{replayb}. For each stored sample vector, $t$ is current time step, $x$ is the collection of state observations at current time step $t$, $r$ is numerical value of reward at current state, $u$ is the action of time step $t$, and finally, the $x_{t+1}$ is the state in the next time step $t+1$ as a result of taking action $u_t$ at state $x_t$. Replay buffer makes sure an independent and efficient sampling \cite{replaye}. 

As discussed above, the DDPG framework is a model-free reinforcement learning architecture, which means it does not seek an explicit model for the dynamical system. Instead, we use a value function network, also called Critic network $Q(s,a|{\theta^{Q}})$, to approximate the result from a state-action pair. The result of a trained Critic network estimates the expected future reward by taking action $u_t$ at state $x_t$. If we take the gradient of the change in the updated reward, we will be able to use that gradient to update our Policy network, also called Actor network. At the end of the training, we obtain an Actor network capable of designing optimal control based on past experience. Since there are two components in the formulation of the gradient of reward in Equation \ref{eqn.DDPG_grad}, we can apply the chain rule to break the process into evaluating gradients from Actor and Critic networks separately. The total gradient is then defined as the mean of the gradient over the sampled mini-batch, where $N$ is the size of the training mini-batch taken from replay buffer $R$, as shown in Equation \ref{eqn.policy}. Silver et al. showed proof in Ref. \cite{silver2014deterministic} that Equation \ref{eqn.policy} is the policy gradient that can guide the DDPG model to search a policy network to yield the maximum expected reward.
\begin{equation}\label{eqn.policy}
    \begin{gathered}
        {\nabla_{\theta^{\mu}}}J \approx \frac{1}{N} \sum_t [\nabla_u Q(x,u|\theta^Q)|_{x=x_t, u=\mu(x_t)} \nabla_{\theta^{\mu}} \mu(x|\theta^\mu)|_{x_t}]
    \end{gathered}
\end{equation}

In this case, dynamical responses from the system are built into the modeling of the Critic network, thus indirectly modeled. The accuracy in predicted reward values is used as the only training criterion in this process, signifying the major difference between the model-free reinforcement learning and the model-based optimal control method such as DKRC. Some of the later observations in different behaviors from model-based vs. model-free comparison have roots stemming from this fundamental difference.

We separate the training of the DDPG into the training of the Actor and Critic networks. 

Actor, or policy network ($\theta^{\mu}$), is a simple neural network with weights $\theta^\mu$, which takes states as input and outputs control based on a trained policy $\mu(x,u|{\theta^{\mu}})$. The Actor is updated using the sampled policy gradient generated from the Critic network, as proposed in Ref. \cite{pga}. 

Critic, or value function network, is a neural network with weights $\theta^Q$, which takes a state-action pair $(x_t,u_t)$ as input. We define a temporal difference error term, $e_t$ (TD error), to track the error between the current output compared to a target Critic network with future reward value using future state-action pair as input in the next time step. The total loss from the Critic network during a mini-batch is defined as $L$ in Equation \ref{eqn.error_loss}, which is based on Q-learning, a widely used off-policy algorithm \cite{qlr}.
\begin{equation}\label{eqn.error_loss}
    \begin{gathered}
        e_t = (r_t + \gamma Q^{'}(x_{t+1}, \mu^{'}(x_{t+1}|{\theta^{\mu^{'}}})| \theta^{Q^{'}})) - Q(x_t,u_t|\theta^{Q})\\
        L = \frac{1}{N} \sum_t{e_t}^2
    \end{gathered}
\end{equation}
In this Critic loss function in Equation \ref{eqn.error_loss}, the terms with prime symbol $(')$ is the Actor/Critic model from the target networks. Target Actor $\mu^{'}$ and target Critic $Q^{'}$ of DDPG are used to sustain a stable computation during the training process. These two target networks are time-delayed copies of the actual Actor/Critic networks in training, which only take a small portion of the new information from the current iteration to update target networks. The update scheme of these two target networks are defined in Equation \ref{eqn.target_nn}, where $\tau$ is Target Network Hyper Parameters Update rate super parameter with a small value ($\tau \ll 1$) to improve the stability of learning.
\begin{equation}\label{eqn.target_nn}
    \begin{gathered}
        {\theta^{Q^{'}}} \leftarrow \tau \theta^Q + (1-\tau)\theta^{Q^{'}}\\
        {\theta^{\mu^{'}}} \leftarrow \tau \theta^{\mu} + (1-\tau)\theta^{\mu^{'}}\\
    \end{gathered}
\end{equation}

To train DDPG, we initialize four networks ($Q(s,a|\theta^Q), \mu(s|\theta^\mu), Q^{'}(s,a|\theta^{Q^{'}}), \mu^{'}(s|\theta^{\mu^{'}})$) and a list buffer ($R$) first, then select an initial action from actor, executing action $a_t$ and observe reward $r_t$ and new state $s_{t+1}$, store $(s_t, a_t, r_t, s_{t+1})$ in the buffer $R$, sampling mini training batch input randomly from R, updating critic by Equation \ref{eqn.error_loss}, updating actor by Equation \ref{eqn.policy}, updating target actor and critic by Equation \ref{eqn.target_nn} until loss value converges.

\bigskip

\section{Experiment Setup}
\begin{figure}[ht]
    \centering
    \includegraphics[width=0.45\textwidth]{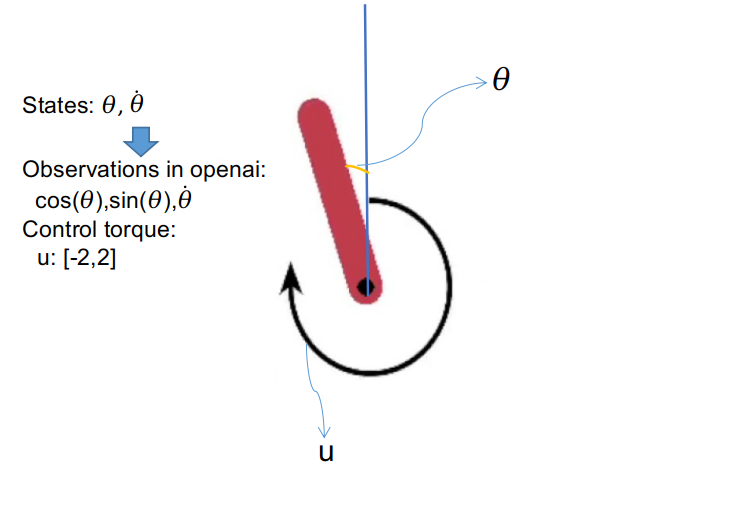}
    \caption{Environment visualization}\label{fig:env_intro}
\end{figure}

To obtain benchmark comparison results, we use the classic 'Pendulum-v0' OpenAI Gym environment \cite{brockman2016openai} to examine the behaviors of controllers built based on DDPG and DKRC for this inverted pendulum problem. The OpenAI Gym is a toolkit designed for developing and comparing reinforcement learning \cite{rl}. The inverted pendulum swing-up problem is a classic problem in the control literature, and the goal of the system is to swing the pendulum up and make it stay upright, as depicted in Figure \ref{fig:env_intro}. Although both the two approaches are data-driven in nature, thus versatile in deployment, We would like to still stick with the classic control problem, which has rich documentation of analytical solution for later comparisons. As will be shown in the next section, the DKRC approach not only successfully finishes the control task using a learned dynamics directly resembling the analytical solution but can also explain the system in lifted dimensions using Hamiltonian Energy Level theorem.
\subsection{Problem set-up}
 A visualization of the simulation environment is shown in Figure \ref{fig:env_intro}. As shown in the picture, the simple system has two states $x = [\theta,\dot{\theta}]$ and one continuous control moment at the joint, which numerically must satisfy $-2\leq u\leq2$. This limit in control magnitude is applied with the intention to add difficulty in designing the control strategy. For the default physical parameter setup, the maximum moment that is allowed for the environment will not be sufficient to raise the inverted pendulum to the upright position in one single move. The control strategy needs to learn how to build momentum by switching moment direction at the right time. The OpenAI Gym defines the observations and control in Equation \ref{eqn.state_space_pen}.\\
\begin{equation}\label{eqn.state_space_pen}
    \begin{gathered}
        \boldsymbol{\chi}=[\cos{\theta}, \sin{\theta}, \dot{\theta}], \quad where\ \theta\in[-\pi,\pi], \dot{\theta}\in[-8,8] \\
        \boldsymbol{U}=[u], \quad where\ u\in[-2,2]
    \end{gathered}
\end{equation}\\
The cost function is designed in Equation \ref{eqn.reward}. The cost function tracks the current states ($\theta$) and control input ($u$). The environment also limits the rotational speed of the up-swing motion, by including a $\dot{\theta}^2$ term in the cost function. By minimizing this cost function, we ask for the minimum control input to ensure the pendulum ends at the upright position with minimum kinetic energy levels. During the comparison, we are using the Equation \ref{eqn.reward} directly in the controller design for the DKRC, whereas we negate cost function and transforms it into a negative reward function, where 0 is the highest reward for the DDPG training. By using this simple simulation environment, we ensure the two approaches are being compared on the same basis with the same reward/cost function definition.
\begin{equation}\label{eqn.reward}
    \begin{gathered}
        reward = \theta^2 + 0.1\dot{\theta}^2 + 0.001u^2
    \end{gathered}
\end{equation}
The dynamical system of the swing-up pendulum can be analytically solved by Euler–Lagrange method \cite{ele}, with its governing equation shown in Equation \ref{eqn.lagrange}, where $\theta = 0$ is the upright position, $g$ is the gravitational acceleration, $m$ is the mass of the pendulum, $l$ is the length of the pendulum.
\begin{equation}\label{eqn.lagrange}
    \begin{gathered}
        \ddot\theta = -\frac{3g}{2l}\sin(\theta + \pi) + \frac{3}{ml^2}u
    \end{gathered}
\end{equation}
By converting the second order ordinary differential equation(ODE) in Equation \ref{eqn.lagrange} to the first order ODE, we can get Equation \ref{eqn.ODE}.
\begin{equation}\label{eqn.ODE}
    \begin{gathered}
        \frac{d}{dt}\begin{bmatrix}
        \cos\theta\\
        \sin\theta\\
        \dot\theta \end{bmatrix} = \begin{bmatrix}
        0 & -1 & 0\\
        1 & 0 & 0\\
        0 & \frac{3g}{2l} & 0
        \end{bmatrix}\begin{bmatrix}
        \cos\theta\\
        \sin\theta\\
        \dot\theta \end{bmatrix}+
        \begin{bmatrix}
        0\\
        0\\
        \frac{3}{ml^2} \end{bmatrix}u\\
    \end{gathered}
\end{equation}
We use the default physical parameters defined in the OpenAI Gym, i.e., $m=1$, $l=1$, $g=10.0$, $dt=0.05$. We discretize the model using zero order hold (ZOH) method \cite{zoh} with a sampling period $dt$. As a result, we can achieve the linearized governing Equation \ref{eqn.zoh} with observed states $y_t$. The numerical values for the hyperparameters used in the default problem setting are also listed in Equation \ref{eqn.zoh_mat}.
\begin{equation}\label{eqn.zoh}
    \begin{gathered}
        x_{t+1} = {A_d}{x_t} + {B_d}{u_t}\\
        y_t = C{x_t}
    \end{gathered}
\end{equation}
with\\
\begin{equation}\label{eqn.zoh_mat}
    \begin{gathered}
        A_d = \begin{bmatrix}
        0.9988 & -0.04998 & 0\\
        0.04998 & 0.9988 & 0.05\\
        0.01875 & 0.7497 & 1
        \end{bmatrix}\
        B_d = \begin{bmatrix}
        0\\
        0\\
        0.15 \end{bmatrix}\\
        C = \begin{bmatrix}
        1 & 0 & 0\\
        0 & 1 & 0\\
        0 & 0 & 1 \end{bmatrix}
    \end{gathered}
\end{equation}
We use $A_d$ and $B_d$ as the benchmark of learned dynamic models comparison between DKRC and Euler-Lagrange Linearization in later sections.

\subsection{Parameters of DDPG and DKRC}
The following Table (\ref{my-label}) is the parameters we use to obtain DDPG and DKRC solutions. Both two methods are trained on NVIDIA Tesla V-100 GPUs on an NVIDIA-DGX2 supercomputer cluster.
\begin{table}[ht]
\centering
    \caption{Parameters of different methods}
\label{my-label}
\begin{tabular}{ |c|c|c|c| } 
\hline
Method & Parameter & Value \\
\hline
 {}
& {Buffer size} & 1e6 \\ 
{}   & {Batch size} & 64 \\ 
{}   & {$\gamma$ (Discount factor)} & 0.9 \\ 
DDPG   & {$\tau$ (Target Network Update rate)} & 0.001 \\
{}   & {Target \& Actor learning rate} & 1e-3 \\
{}   & {Target \& Critic learning rate} & 1e-2 \\
{}   & {Training epochs} & 5e4 \\
\hline
DKRC 
& {Lift dimension} & 8 \\ 
{}   & {Training epochs} & 70 \\ 
\hline
\end{tabular}
\end{table}

The result of the DDPG solution is an Actor neural network with optimal parameters for the nonlinear pendulum system - $\mu(x;\theta^{*})$.

The result of the DKRC solution is identity matrices $A_{lift}$, $B_{lift}$, $C$ of the lifted space, and a lift neural network $\psi_{N}(x;\theta)$ for observations of the unknown dynamical system.

\bigskip
\section{Results and Comparison}
 
\subsection{Control Strategies of DDPG and DKRC}
\begin{figure}[!htbp]
\centering
\includegraphics[width=0.4\textwidth]{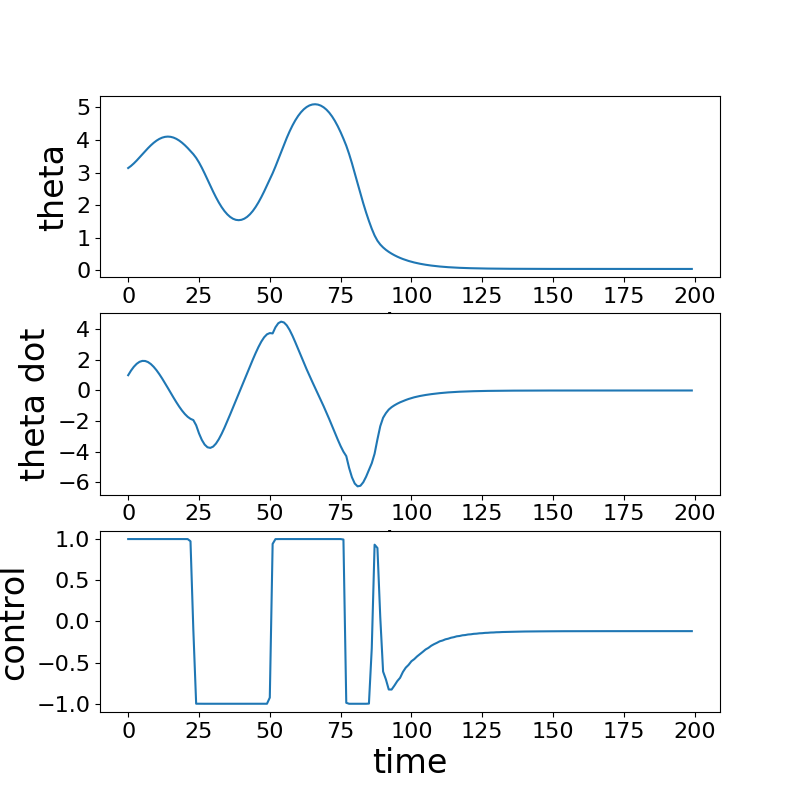}
\includegraphics[width=0.4\textwidth]{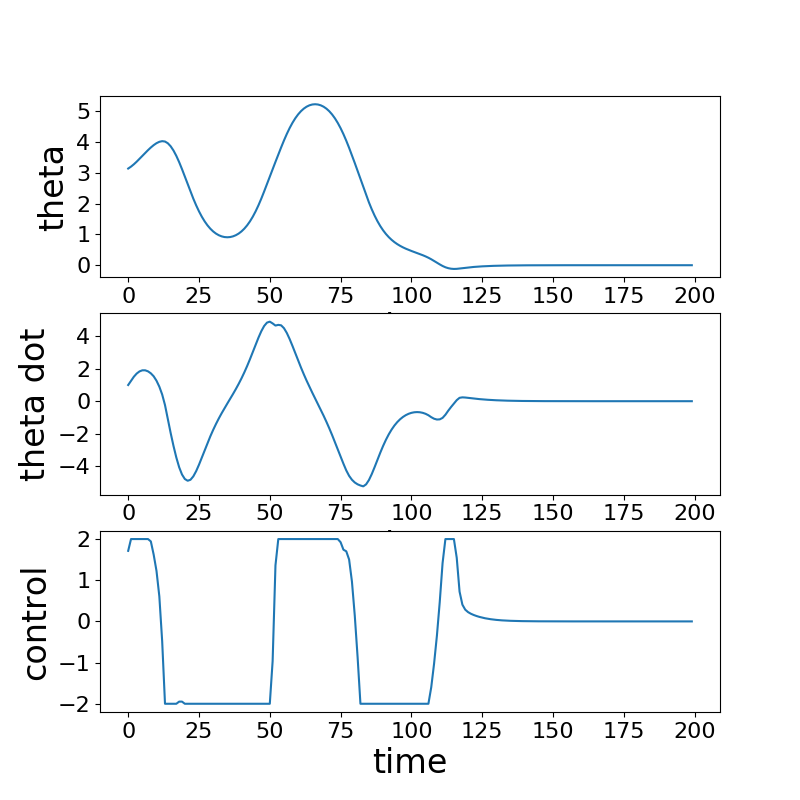}
\caption{Trajectories of Swinging Pendulum implementing DDPG model (left) vs. DKRC model (right) initialized at the lowest position}\label{fig:traj_lowes}
\end{figure}

\begin{figure}[!htbp]
\centering
\includegraphics[width=0.4\textwidth]{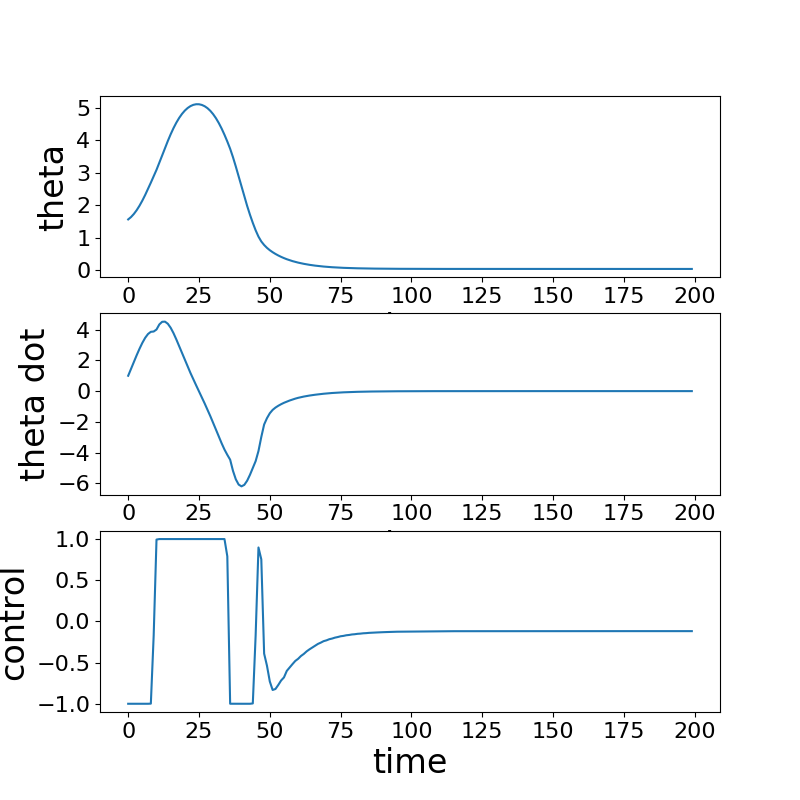}
\includegraphics[width=0.4\textwidth]{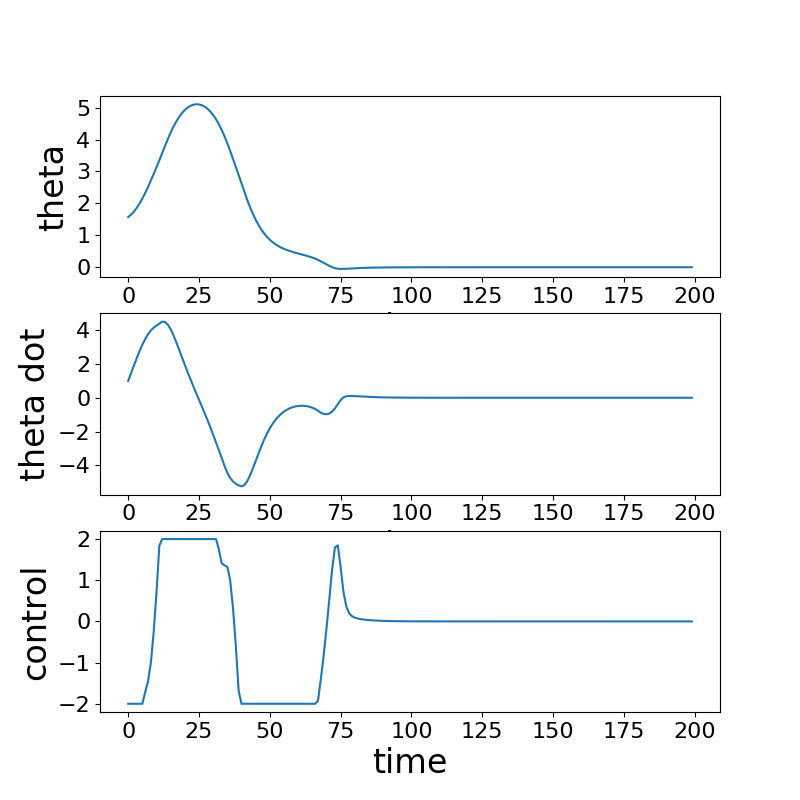}
\caption{Trajectories of Swinging Pendulum implementing DDPG model (left) vs. DKRC model (right) initialized at the left horizontal position}\label{fig:traj_lh}
\end{figure}

\begin{figure}[!htbp]
\centering
\includegraphics[width=0.4\textwidth]{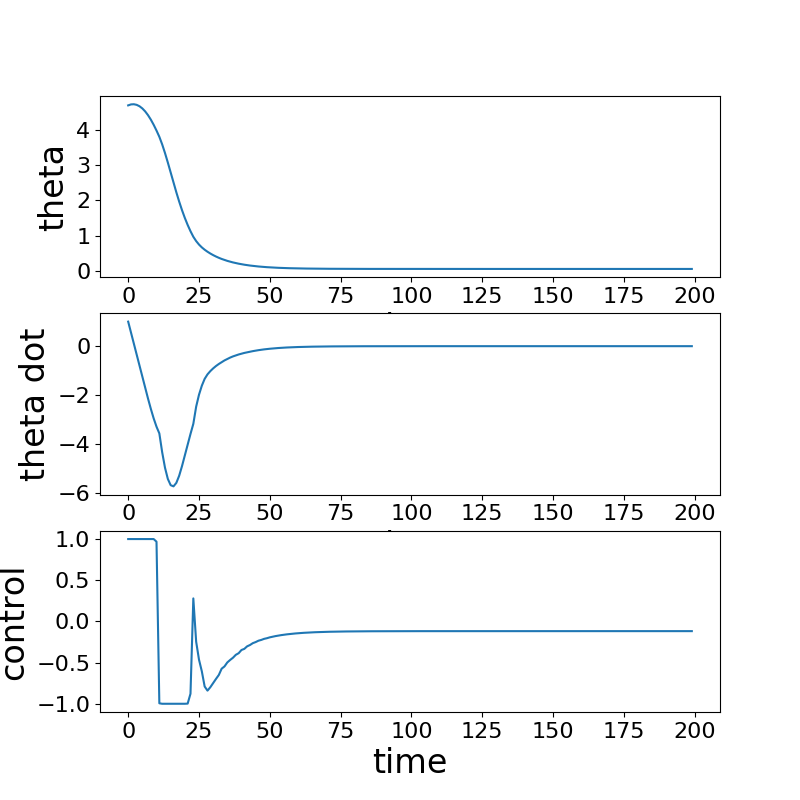}
\includegraphics[width=0.4\textwidth]{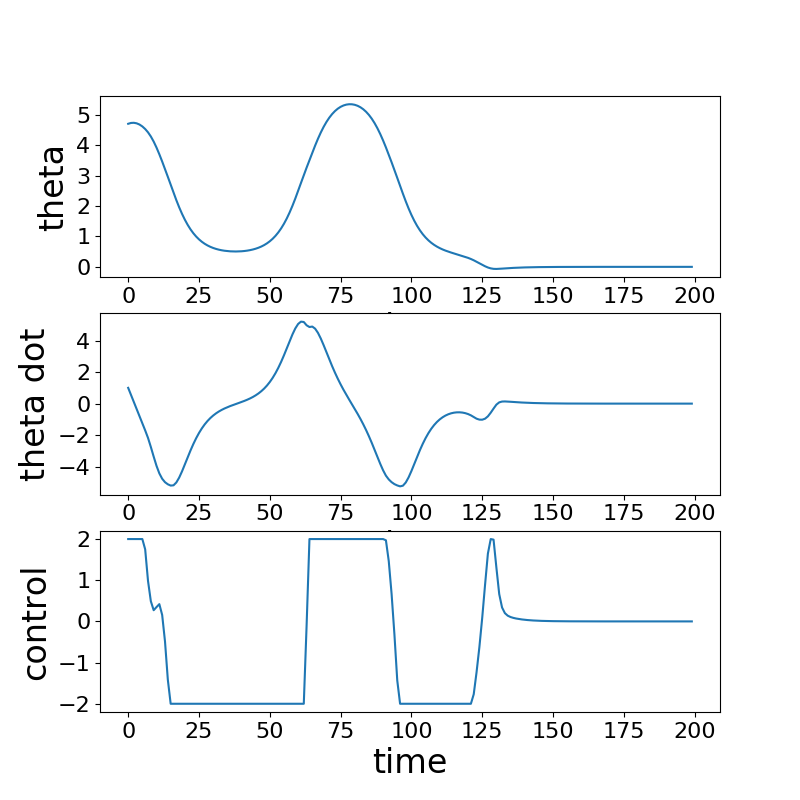}
\caption{Trajectories of Swinging Pendulum implementing DDPG model (left) vs. DKRC model (right) initialized at the right horizontal position}\label{fig:traj_rh}
\end{figure}

\begin{figure}[!htbp]
\centering
\includegraphics[width=0.4\textwidth]{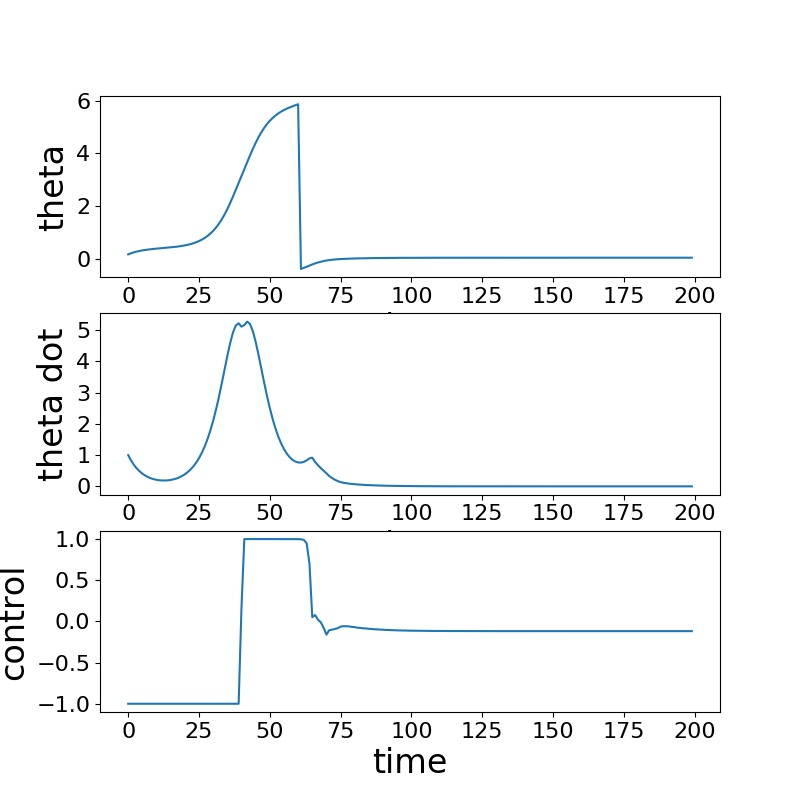}
\includegraphics[width=0.4\textwidth]{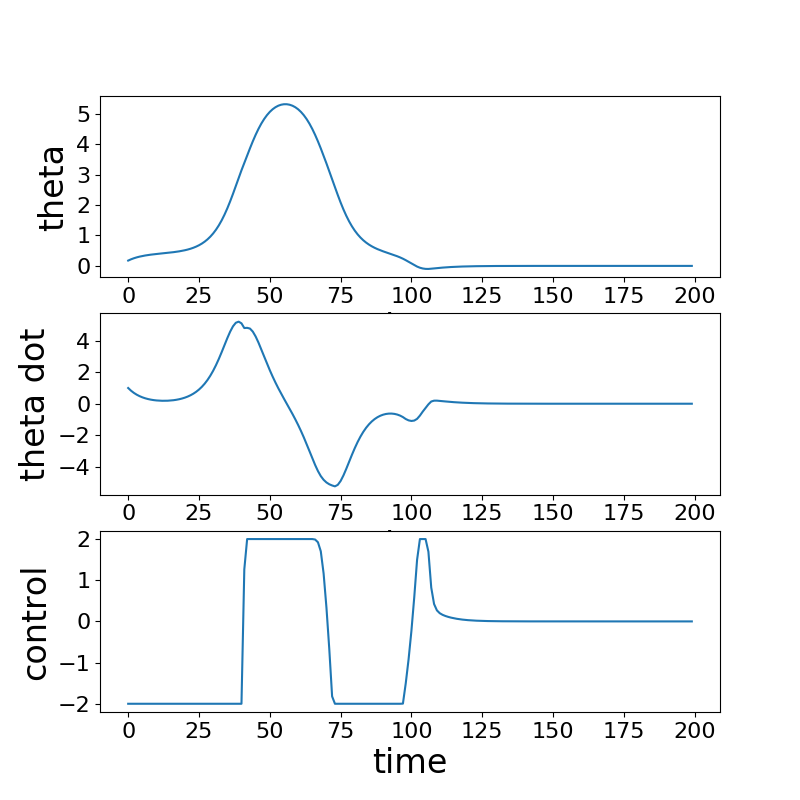}
\caption{Trajectories of Swinging Pendulum implementing DDPG model (left) vs. DKRC model (right) initialized at the close left upright position}\label{fig:traj_lu}
\end{figure}

We present results from the DKRC vs DDPG by specifying five initialization configuration for the problem. We have choices of defining the starting position and also the initial disturbance in the form of starting angular velocity of the pendulum. In this study, we initialize the pendulum at five different positions:\ $\theta_0 = \pi$ (lowest position), \ $\theta_0 = \frac{\pi}{2}$ (left horizontal position), \ $\theta_0 = -\frac{\pi}{2}$ (right horizontal position), $\theta_0 =  \pm \frac{\pi}{18}$ (close to upright position). The initial angular velocity of pendulum at different initial positions is $\dot{\theta_0} = 1rad/s$ (clockwise). For better visualization, we map the angle from $[-\pi, \pi]$ to $[0, 2\pi]$, where the upright position (goal position) is always achieved at $\theta=0$ and $\dot{\theta} = 0$. The result is shown in Figure \ref{fig:traj_lowes}$-$ \ref{fig:traj_lu}.

\begin{figure}[!htbp]
\centering
\includegraphics[width=0.4\textwidth]{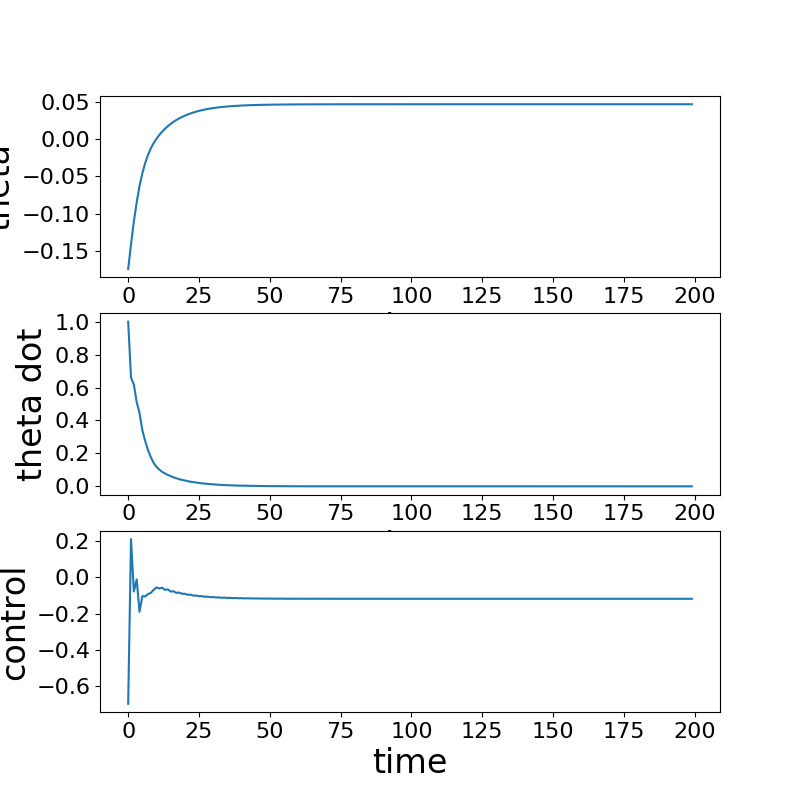}
\includegraphics[width=0.4\textwidth]{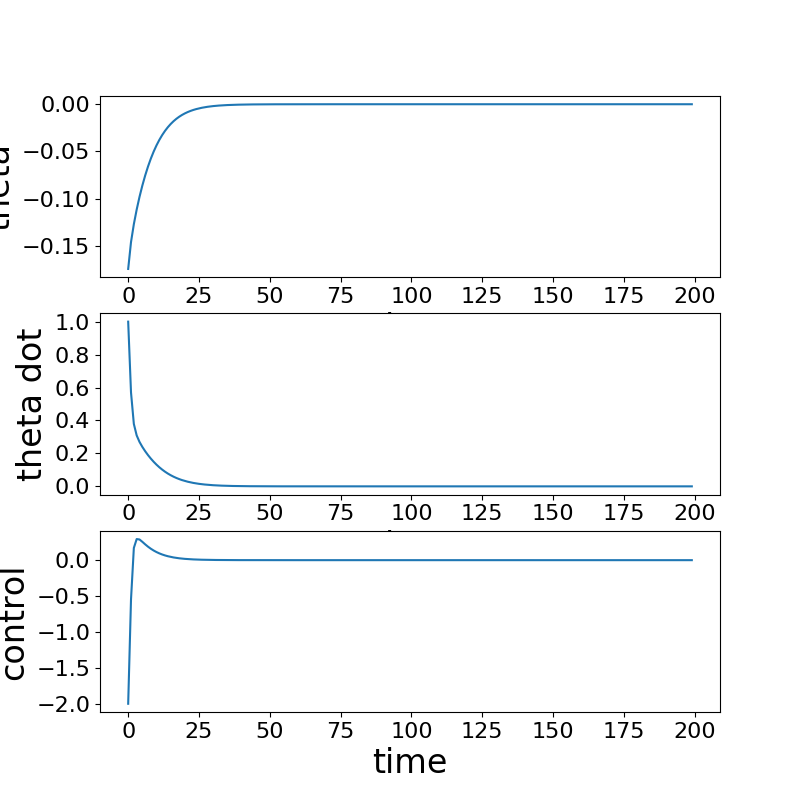}
\caption{Trajectories of Swinging Pendulum implementing DDPG model (left) vs. DKRC model (right) initialized at the close right upright position}\label{fig:traj_ru}
\end{figure}

Figure \ref{fig:traj_lowes}$-$ \ref{fig:traj_ru} show that DDPG and DKRC have similar control strategies in most initial positions. DDPG needs less time to arrive at the goal position than DKRC when the pendulum is initialized on the right side. It tends to use smaller control torque as a result of that DDPG has constraints term for control torque. Still, DDPG never succeeds in getting an absolute upright position, i.e., at the final position, a non-zero control input is always required to sustain a small displacement away from the goal position. On the contrary, DKRC can achieve a precise goal position with much less training time than DDPG. Once a proper dynamical system model can be learned directly from data, it makes more sense to execute control using model-based controller design such as MPC.

\begin{figure}[ht]
\centering
\includegraphics[width=0.42\textwidth]{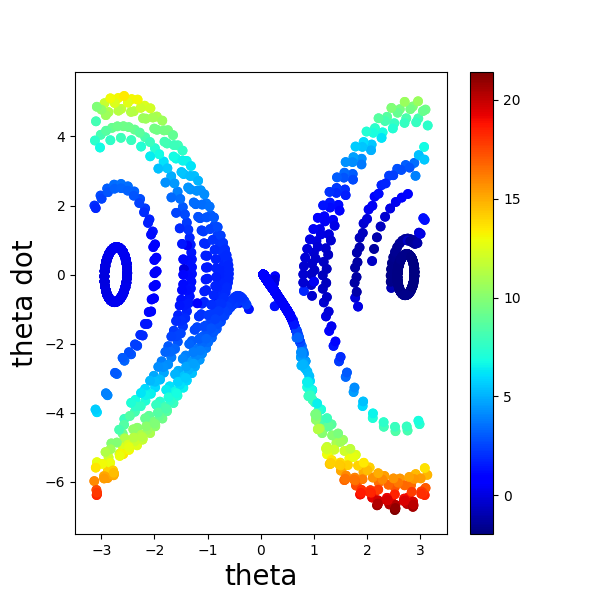}
\includegraphics[width=0.42\textwidth]{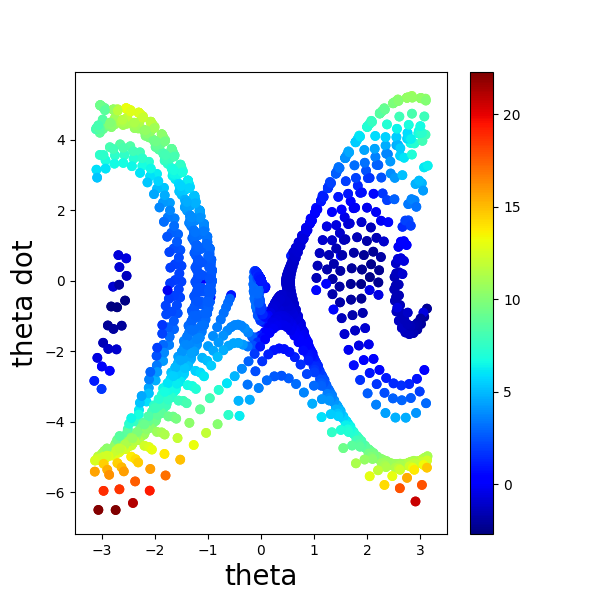}
\caption{50 pendulum games data recorded using DDPG model (left) vs. DKRC model (right), color mapped by energy}\label{fig:deployment}
\end{figure}

Another way to show the differences in control strategies is by plotting the measured trajectories during repeated tests. In Figure \ref{fig:deployment}, we test the pendulum game for 50 games with a total of 10000 time-steps utilizing both methods (DDPG \& DKRC) solutions. 
In this comparison, we plot the measurements of $\dot{\theta}$ vs. $\theta$ on a 2D basis, colored by the magnitude of the cost function defined in Equation \ref{eqn.reward}. The goal is to arrive at the goal position ($\dot{\theta}=0$ and $\theta=0$, the center of each plot) as quickly as possible. Figure \ref{fig:deployment} shows that DDPG tends to drive the states into "pre-designed" patterns, and execute similar control strategies for the 50 games. Therefore the data points on the left subplot appear to be less than the one on the right. DKRC, on the other hand, tends to exhibit different control behavior due to the local replanning using MPC. The result of the local replanning is that it generates multiple trajectories solving the 50 games with different random initializations. During this comparison, we illustrate that DDPG is indeed deterministic, which is a good indicator of the system's reliability. However, we want to point out that the robustness of the system will also benefit from a local replanner available under different initializations or disturbances since the "pre-designed" patterns are learned from past experience, therefore, cannot guarantee a viable solution for unforeseeable situations when we move onto more complex systems.

\begin{figure}[ht]
\centering
\includegraphics[width=0.49\textwidth]{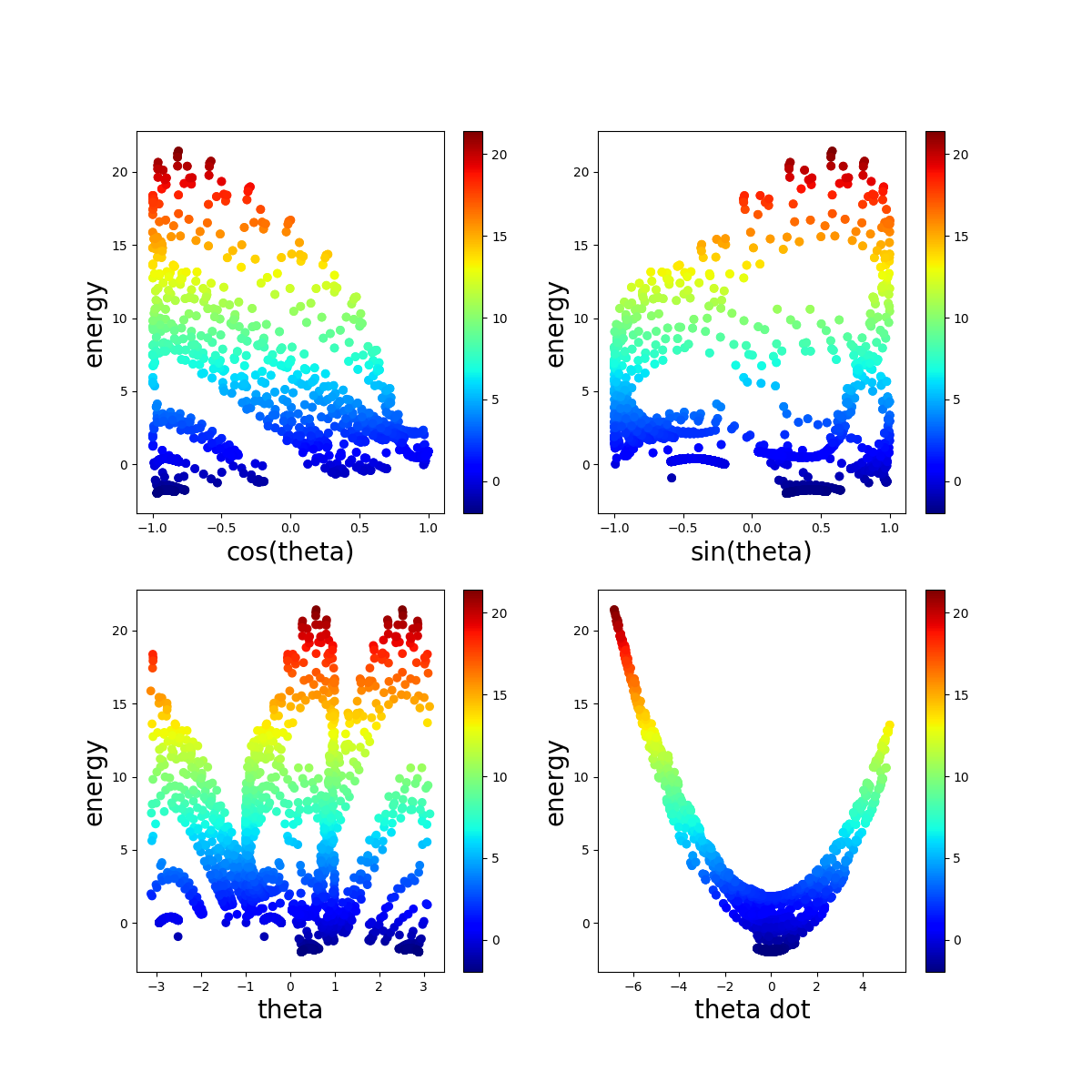}
\includegraphics[width=0.49\textwidth]{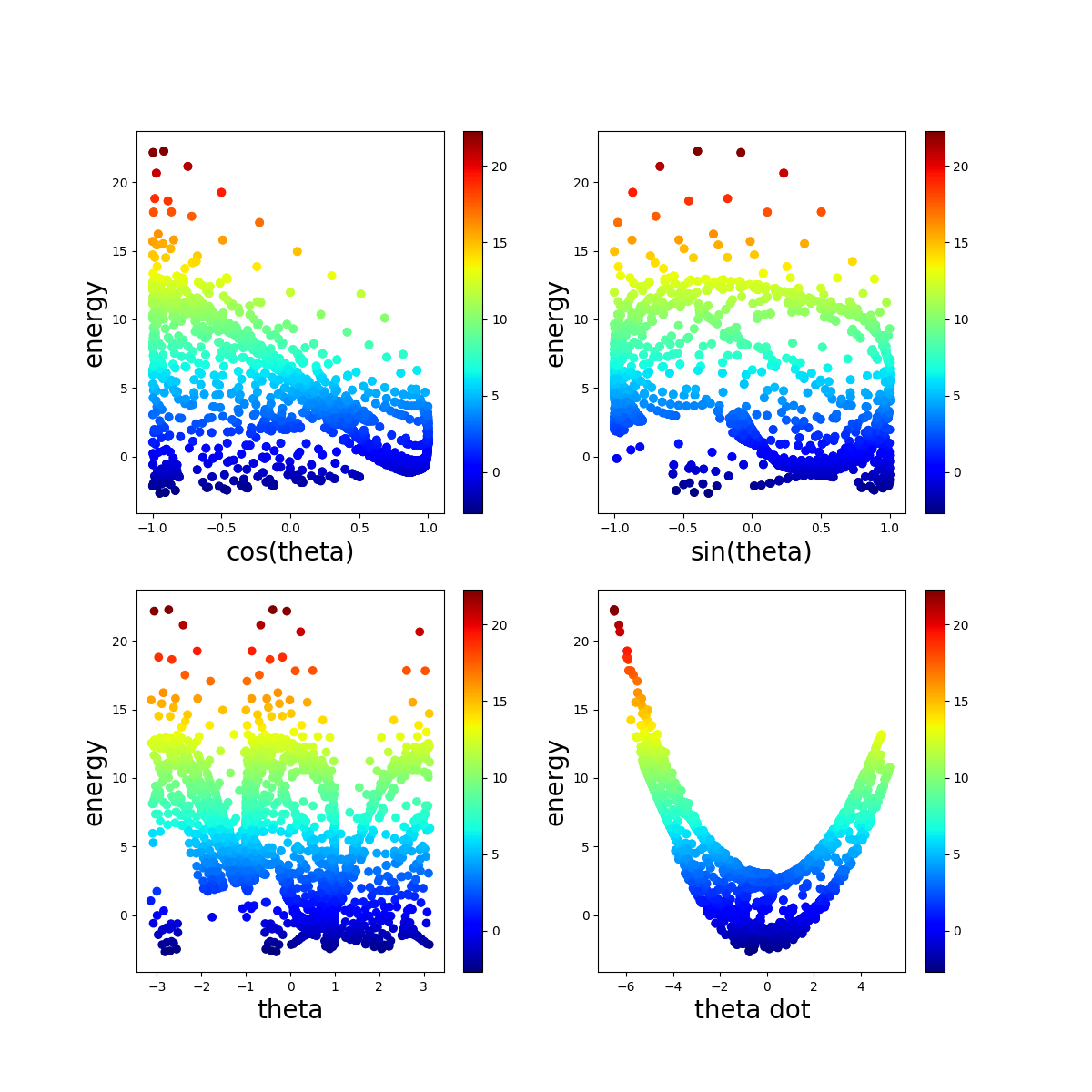}
\caption{DDPG(left) vs. DKRC(right): Recorded Observations \& Energy}\label{fig:pendulum_energy}
\end{figure}

By transforming the observed states, we can also show the relationship between the designed control and the energy levels in the system. Consequently, our control goal is to achieve the lowest energy level of the system in terms of lowest magnitudes in both kinetic energy and potential energy. In Figure \ref{fig:pendulum_energy}, we present the Hamiltonian energy level plots with respect to the measured states ($\theta$, $cos(\theta)$, $sin(\theta)$, and $\dot{\theta}$) for the forced frictionless pendulum system, each colored by the same energy level scale. The energy term used in forced pendulum is defined as $\frac{1}{2}\dot{\theta}^2+cos(\theta)+u$.
Figure \ref{fig:pendulum_energy} shows that DKRC's trajectories are concentrated in lower energy areas compared to DDPG, which means it intends to minimize the energy directly. This behavior also resembles many design strategies used in classic energy-based controller design approaches.

\subsection{Control Visualization using Decoder Neural Network}
\begin{figure}[ht]
\centering
\includegraphics[width=0.5\textwidth]{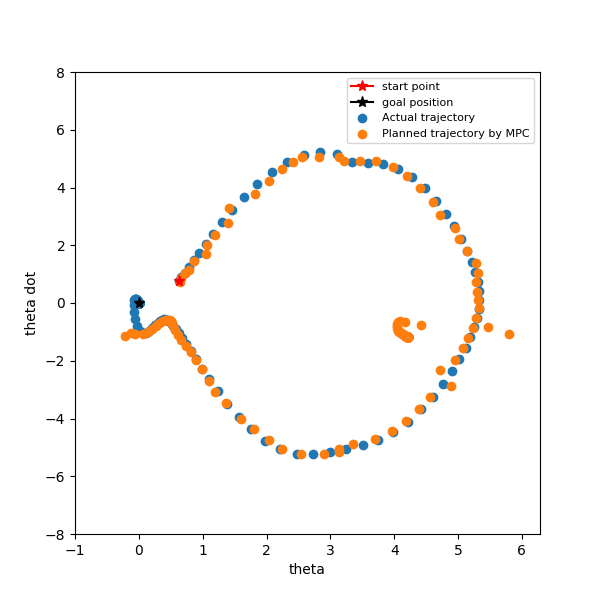}
\caption{DKRC's planned trajectory and actually executed trajectory in Inverted pendulum}\label{fig:mpctraj}
\end{figure}

Another benefit of having a model-based controller design is the interpretability of the system. We can preview the design trajectory since our MPC controller implements the receding horizon control. By deploying one pendulum game, we obtain one planned trajectory for the linear system in the lifted dimension space. To get the comparison between the planned trajectory and the measured trajectory in the state space, we utilize the decoder neural network obtained during the training to map the trajectory in higher-dimensional space (8-dimension for inverted pendulum) to the lower-dimensional space (2-dimension). The comparison between such recovered trajectory planning and the actual trajectory is presented in Figure \ref{fig:mpctraj}. In this figure, we again mapped the $\theta$ in the range of $0$ to $2\pi$ for visualization purposes, whereas the goal position is still at the $\dot{\theta}=0$ and $\theta=0$ position.

In this example case, we initialize the pendulum close to the goal position but give it a moderate initial angular velocity pointing in the opposite direction to the goal position. The DKRC can plan a simple trajectory with continuous control. During the process, we executed MPC multiple times and used feedback measurements to improve the design trajectory. The planned trajectory is being closely followed except certain locations close to $0$ and $\pi$ position. The outlier behavior comes from the neural network treatment for the discontinuity and does not pollute the efficient control for the entire problem. It is worth noticing that, to arrive at this result, we learn the unknown nonlinear dynamics using a purely data-driven approach, and we have to go through an encoding-decoding process to recover the planned trajectory. It is promising to state that the use of Koopman representation for nonlinear control can help with system interpretability, which is currently an active research area.

\begin{figure}[ht]
\centering
\includegraphics[width=0.46\textwidth]{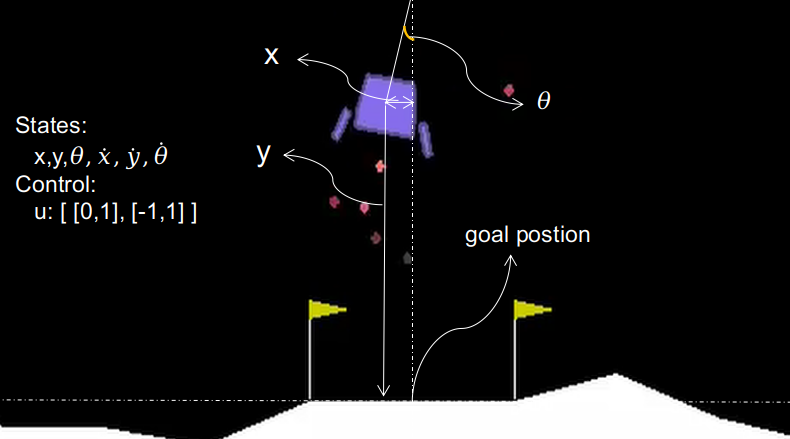}
\caption{Lunar Lander Environment}\label{fig:envll}
\end{figure}

\begin{figure}[ht]
\centering
\includegraphics[width=0.6\textwidth]{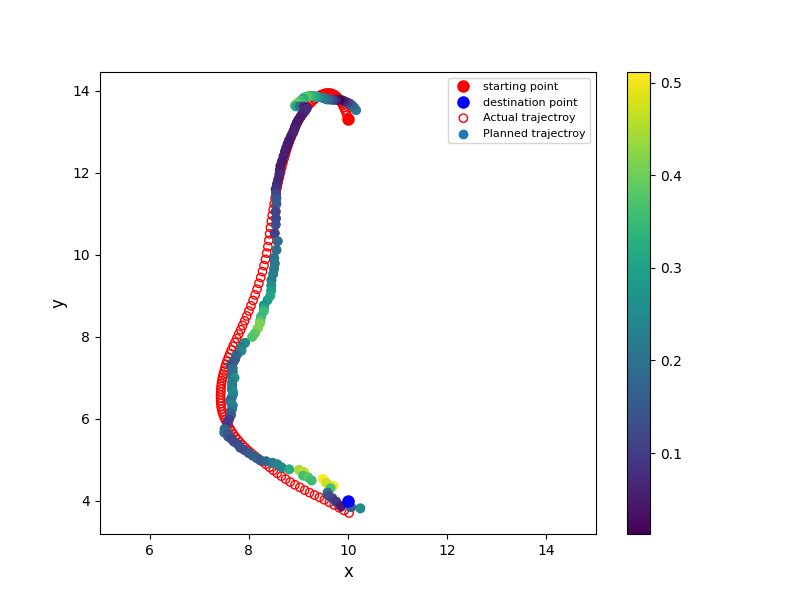}
\caption{DKRC's planned trajectory and actual executed trajectory in Lunar Lander, color mapped by the distance between the planned point and the executed point}\label{fig:llmpctraj}
\end{figure}

To verify whether DKRC is valid in a more complex environment than Inverted pendulum, we also deploy it in the Lunar Lander - continuous control environment of OpenAI Gym. A simple explanation of 'Lunar Lander' is exhibited in Figure \ref{fig:envll}. The control goal is to guide the lunar lander to arrive at the landing zone as smoothly as possible \cite{Han:2020f}. The system is also unknown and must be learned from data. We implement the DKRC framework and use MPC for trajectory planning, and the result is shown in Figure \ref{fig:llmpctraj}.

The actual trajectory measured in state-space is shown in hollow red circles. The planned trajectory is colored by the distance away from the originally planned location. It is evident that in the region where the originally planned location is in the immediate vicinity (dark blue color), the actual trajectory is following the plan very precisely. We implement a finite-horizon control during each MPC planning phase. We plan and execute control with several steps beyond the current state as displayed in lighter green color. The actual trajectory slowly deviates from the planned trajectory when it starts to drift away from the originally planned location. This behavior is expected since we are relying on open-loop control during those finite horizon plannings. The actual trajectory and the projected trajectory merge again once the next round of the MPC control is executed.

In this figure, we demonstrate that the deviation from the planned trajectory and the measured trajectory is from the open-loop planning, rather than from the error introduced while passing the state inputs through the encoder-decoding neural network. The proposed structure is capable of recovering the designed strategy in higher-dimensional space and improving the system's interpretability.

\subsection{Robustness Comparison}
\begin{figure}[ht]
\centering
\includegraphics[width=0.48\textwidth]{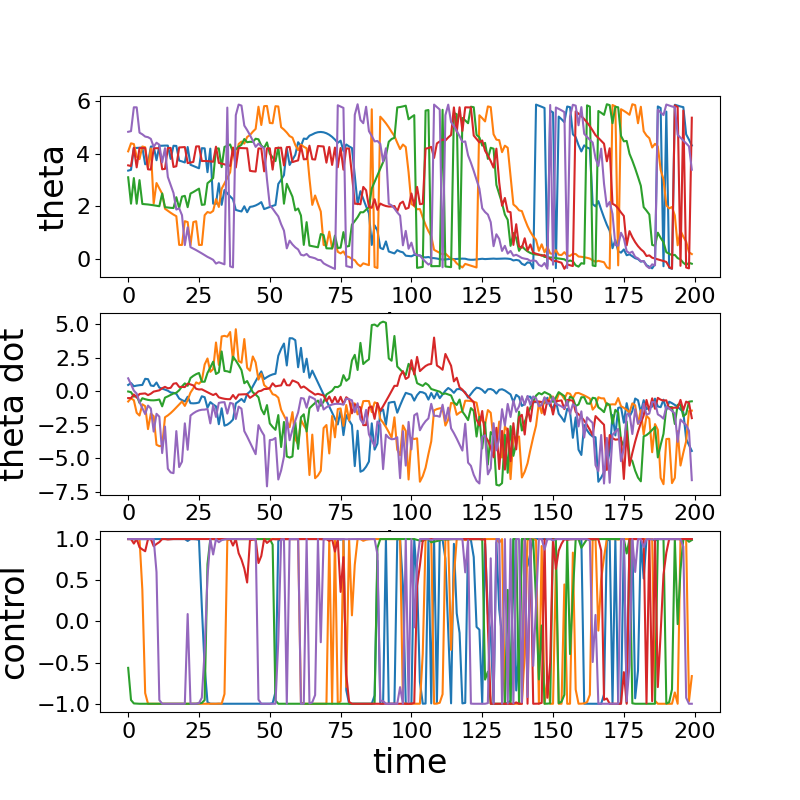}
\includegraphics[width=0.48\textwidth]{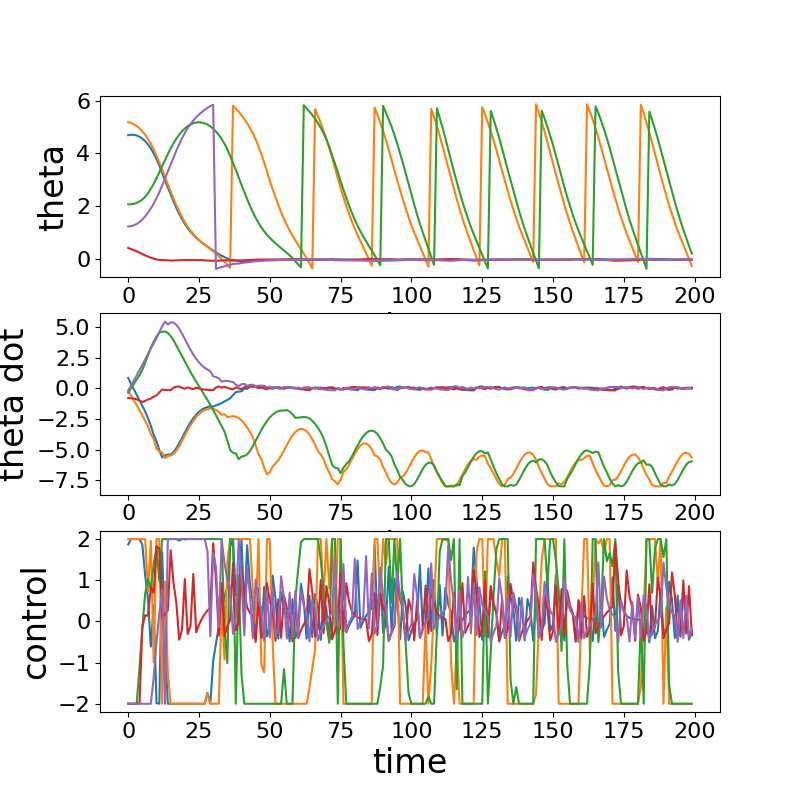}
\caption{Five pendulum games with input states noise using DDPG (left) vs. DKRC (right) solutions}\label{fig:stable}
\end{figure}

To compare the robustness of these two methods, we introduce noises to the state measurements and observe the control outcome from DDPG and DKRC. The noise is designed as multiplying the states with a noise ratio, which is randomly selected from range $[0.6, 1]$ during deployment. In this test, we assume the learned dynamics from DDPG and DKRC are not affected by the noise; only the observations during deployment are affected, representing a cyber-security attack during the operation phase. The new state inputs would be $x = x*noise$ in this scenario. The result for five repeat games is shown in Figure \ref{fig:stable}.
 Each line with the same color in the subplot represents measurements from a single game. Even with a high noise ratio, DKRC can succeed in the control task for three out of five games, whereas DDPG fails every game in Figure \ref{fig:stable}. In this test, we demonstrate that the DKRC is capable of designing a control strategy based on noisy data and continuously adjust the control based on the feedback control loop so that it is still robust in a noisy environment. The robustness of DKRC is another advantage compared to a neural-network-based control system, which relies heavily on state measurement accuracy.

\subsection{Learned dynamics of DKRC compared to Euler-Lagrange analytical solution}
 To illustrate the validity of the learned dynamics using DKRC, we present benchmark comparisons between the proposed DKRC framework and the Euler-Lagrange method's analytical solutions. As previously shown in Equation \ref{eqn.zoh} and \ref{eqn.zoh_mat}, we have obtained the identity matrices for the linearized system using ZOH method. We are making the same assumption for the linearized system in the lifted-dimensional space by the DKRC method. For comparison with different dimension embeddings, we pick the matrices corresponding to the top $n$ eigenvalues from the $A$ and $B$ obtained through DKRC. For the inverted pendulum problem, we collect $K$ time-step data (only need $K=2000$ data points) to obtain Koopman representation of the system. The resultant dimension is $N=8$ ($K \gg {N}$). The learned dynamics $A, B$ of the lifted linear system $\psi_{N}(x|\theta)$ are shown in the following matrices\\
\\
$A_{DRKC} =
    \scriptsize{\begin{bmatrix}
                \begin{array}{ccc|ccccc}
    1.01 & -0.06 & -0.01 & -0.08 & -0.08 & 0.06 & -0.01 & -0.01 \\
    0.00 & 0.98 & -0.04 & 0.02 & 0.06 & -0.07 & 0.01 & 0.02 \\
    0.06 & 0.05 & 0.93 & 0.06 & 0.00 & 0.00 & -0.03 & 0.02\\
    \hline
    0.03 & -0.00 & -0.01 & 0.97 & -0.04 & 0.05 & -0.01 & -0.01\\
    0.11 & 0.02 & -0.03 & 0.01 & 0.98 & 0.05 & -0.07 & 0.01\\
    0.04 & 0.05 & 0.01 & 0.02 & 0.01 & 1.01 & -0.05 & 0.04\\
    0.01 & 0.014 & 0.02 & 0.01 & 0.08 & -0.04 & 0.96 & 0.04\\
    -0.00 & -0.01 & -0.02 & -0.00 & 0.08 & -0.08 & -0.00 & 1.00
    \end{array}
    \end{bmatrix}} \ 
B_{DRKC} = \scriptsize{\begin{bmatrix}
    0.00014\\
    0.00018\\
    -0.00024\\
    0.00017\\
    0.00021\\
    0.00038\\
    0.00008\\
    -0.0001 \end{bmatrix}}$\\ \\
 To measure the similarity of $A_d$ (in Equation \ref{eqn.zoh_mat}), $A_{DDPG}$, $A_{DKRC}$ we achieved, we adopt the Pearson correlation coefficient(PCC) method \cite{pcc} in Equation \ref{eqn.pcc}, where matrices with bar operator, e.g. $\bar{A}$, is the sample mean of that matrix. The result $r(A,B)\in[0,1]$ represents the correlation between the two matrices. Two matrices ($A$ and $B$) are more similar when $r$ is closer to 1.\\
     \begin{equation}\label{eqn.pcc}
        \begin{gathered}
            r(A,B) = \frac{\sum_m \sum_n(A_{mn}-\bar{A})(B_{mn}-\bar{B})}{\sqrt{(\sum_m \sum_n(A_{mn}-\bar{A})^2)(\sum_m \sum_n(B_{mn}-\bar{B})^2)}}
        \end{gathered}
    \end{equation}\\
To compare the similarity of these two matrices with different dimensions, we extract the left top $3\times3$ part of the $A_{DKRC}$. We achieve a correlation score of $r(A_{d}, A_{DKRC})=0.8926$, which means the lifted linear system of DKRC is very similar to the analytical system model solved by Euler-Lagrange method. This high correlation score indicates that the data-driven Koopman representation of the system can reveal the intrinsic dynamics purely based on data samples.

\begin{figure}[ht]
\centering
\includegraphics[width=0.48\textwidth]{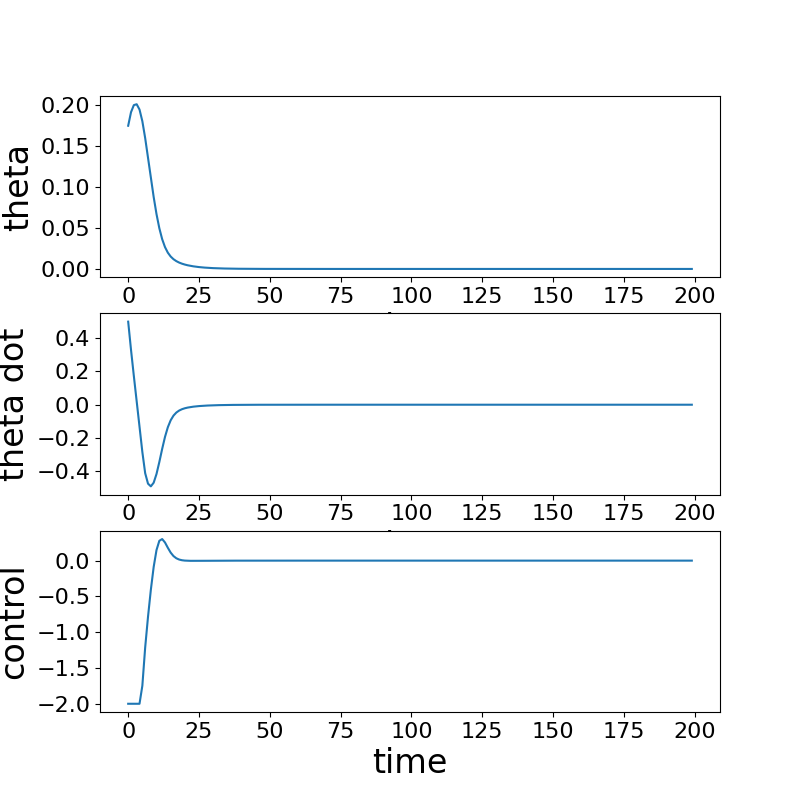}
\includegraphics[width=0.48\textwidth]{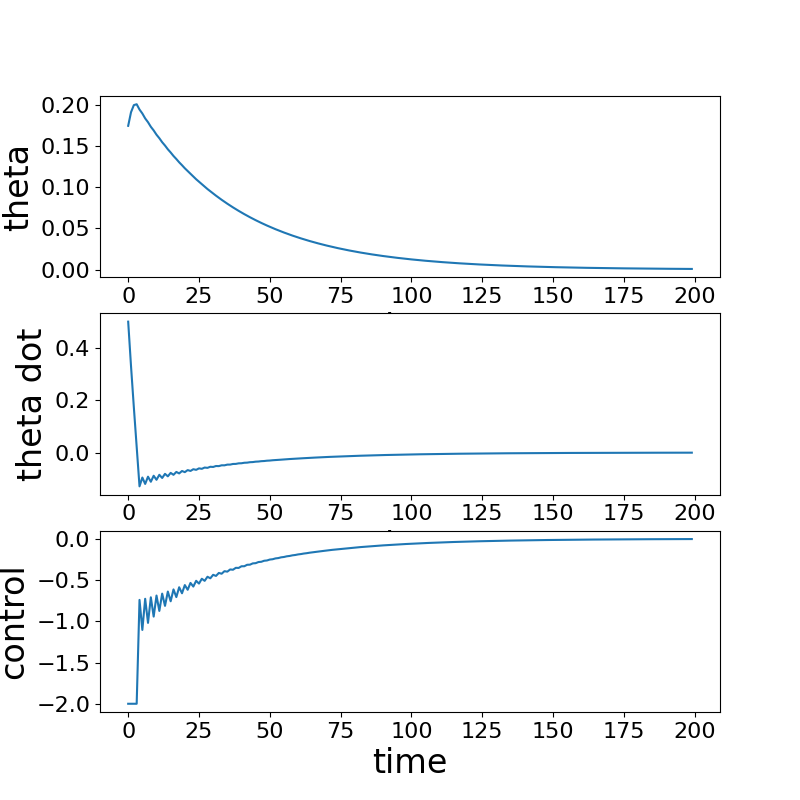}
\caption{One pendulum game using DKRC model (left) vs. Euler-Lagrange method (right); Both tests initialized at $\theta_0 = \frac{\pi}{18}$, $\dot{\theta_0} = 0.5$}\label{fig:lec}
\end{figure}

On a separate note, we do not expect the two results are exactly the same since by lifting the system to a higher-dimensional space, we have more neurons in DKRC neural networks to store the system information that was not included in the previous comparison. In addition, we also deploy the linearization model obtained by the Euler-Lagrange method with MPC to the same OpenAI Gym environment to examine the effectiveness by direct linearization without lifting. Unlike DKRC or DDPG, which can work for any arbitrary initial configuration, control designed by the Euler-Lagrange method only works when the pendulum's initial position is between $[-23.4^{\circ}, 23.4^{\circ}]$ with small initial angular velocity disturbance. A sample comparison between Euler-Lagrange MPC and DKRC is exhibited in Figure \ref{fig:lec}.

As shown in Figure \ref{fig:lec}, DKRC only spends around $20\%$ time of Euler-Lagrange linearization method to make pendulum converge to the upright position.

\section{Conclusions}
We provide a systematic discussion of two different data-driven control frameworks: Deep Deterministic Policy Gradient (DDPG) and Deep Learning of Koopman Representation for Control (DKRC). Both the two methods are data-driven and adopt neural networks as central architecture, but the controller design is model-free for DDPG, whereas DKRC utilizes a model-based approach. Our experiments in a simple swing-up pendulum environment demonstrate the different solutions achieved by DKRC and DDPG. The DKRC method can achieve the same control goal as effective as the DDPG method but requires much less training epochs and training samples (70 epochs vs. $5\times10^4$ epochs). Due to the physics model-based nature, DKRC provides better model interpretability and robustness, which are both critical for real-world engineering applications.

 
 

\section*{Acknowledgment}
Y.Han would like to acknowledge research support from ONR award N00014-19-1-2295.

\bigskip
\bibliographystyle{unsrt}
\bibliography{ref_hao}

\end{document}